\documentclass[conference]{IEEEtran}
\IEEEoverridecommandlockouts

\usepackage{amsmath,amssymb}
\usepackage{graphicx}
\usepackage{booktabs}
\usepackage{multirow}
\usepackage{cuted}
\usepackage{capt-of}
\usepackage{algorithm}
\usepackage{algpseudocode}
\usepackage{cite}
\usepackage{url}
\usepackage{microtype}

\title{Balanced Adaptive Prototype Selection for Scalable TabPFN Inference on Large-Scale Tabular Data}

\author{
\IEEEauthorblockN{1\textsuperscript{st} Mahboobe Jadid}
\IEEEauthorblockA{Department of Computer Engineering\\
Ne. C., Islamic Azad University\\
Neyshabur, Iran\\
mahboobe.jadid@iau.ac.ir}
\and
\IEEEauthorblockN{2\textsuperscript{nd} Melika Rezaye Garkani}
\IEEEauthorblockA{Department of Computer Engineering\\
Ne. C., Islamic Azad University\\
Neyshabur, Iran\\
melika.rezaegarkani@iau.ac.ir}
\and
\IEEEauthorblockN{3\textsuperscript{rd} Ali Mousavi}
\IEEEauthorblockA{Department of Computer Engineering\\
Ne. C., Islamic Azad University\\
Neyshabur, Iran\\
mousavi@iau.ac.ir}
}

\begin{document}
\maketitle

\begin{abstract}
Pretrained tabular foundation models have demonstrated strong predictive capability; however, their application to large-scale datasets remains constrained by the limited inference context. This paper introduces Balanced Adaptive Prototype Selection (BAPS), a framework for constructing compact, information-preserving contexts for scalable TabPFN inference. Without modifying or retraining the pretrained model, BAPS jointly preserves representative structure, informative decision boundaries, local density, class balance, and feature-space diversity. Experiments on the million-row HIGGS and SUSY datasets show that 512 prototypes retain strong predictive performance and reliable calibration, corresponding to an approximately 1,953-fold context compression. All experiments were conducted on an Intel Core i7 CPU with 16 GB RAM and no GPU acceleration. These findings establish effective context construction as a practical mechanism for extending pretrained tabular foundation models to million-scale datasets.
\end{abstract}

\begin{IEEEkeywords}
TabPFN, Tabular Foundation Models, Prototype Selection, Context Construction, Large-Scale Tabular Learning
\end{IEEEkeywords}

\section{Introduction}
Recent advances in pretrained foundation models have transformed machine learning by replacing task-specific optimization with knowledge transfer from large-scale pretraining \cite{vanbreugel2024position}. This paradigm has recently been extended to tabular learning, where models such as TabPFN perform prediction directly from a labeled inference context rather than retraining model parameters for each downstream task \cite{hollmann2023tabpfn}. This capability enables efficient deployment while achieving competitive predictive performance across diverse tabular problems \cite{hollmann2025accurate}.

Despite these advantages, the practical scalability of pretrained tabular foundation models remains fundamentally constrained by the bounded inference context \cite{feuer2024tunetables}. Modern tabular datasets often contain hundreds of thousands or even millions of training instances, whereas only a limited number of examples can be presented to the model during inference \cite{hollmann2025accurate}. Consequently, the effectiveness of TabPFN depends not only on the pretrained model itself, but also on whether the selected context preserves the predictive information contained in the original training data \cite{hollmann2023tabpfn,feuer2024tunetables}. Under severe compression, conventional sampling or prototype reduction may discard informative decision boundaries, minority-class evidence, and locally important structures, substantially degrading inference quality \cite{triguero2012prototype}.

Existing prototype selection and dataset reduction methods primarily optimize computational efficiency through sampling, clustering, or instance selection \cite{bien2011prototype,garcia2008memetic}. Although effective for conventional machine learning algorithms, these methods are not designed for pretrained foundation models, where the objective is information-preserving context construction rather than dataset reduction \cite{hollmann2023tabpfn}. Consequently, reducing the number of training samples alone is insufficient for scalable TabPFN inference \cite{feuer2024tunetables}. A principled mechanism for constructing compact contexts that preserve predictive information is therefore essential.

Motivated by this observation, we propose Balanced Adaptive Prototype Selection (BAPS), an information-preserving context-construction framework for scalable TabPFN inference. Instead of modifying or retraining the pretrained model, BAPS constructs compact inference contexts by jointly preserving representative structure, informative decision boundaries, local density, class balance, and feature-space diversity. By optimizing the information retained within a fixed context budget, BAPS directly addresses the scalability bottleneck of pretrained tabular foundation models while remaining fully compatible with the original TabPFN architecture \cite{hollmann2023tabpfn}. Extensive experiments on the million-row HIGGS and SUSY benchmark datasets demonstrate that BAPS maintains strong predictive performance and reliable calibration using only 512 prototypes, corresponding to an approximately 1,953-fold context compression. These results show that effective context construction is not merely an optimization strategy but a fundamental requirement for scaling pretrained tabular foundation models to million-scale datasets.

The main contributions of this work are summarized as follows.
\begin{itemize}
    \item We identify bounded inference context as the primary scalability bottleneck of pretrained tabular foundation models and formulate scalable context construction as an information-preservation problem.
    \item We propose BAPS, an information-preserving context-construction framework that jointly preserves representative structure, informative decision boundaries, local density, class balance, and feature-space diversity without modifying or retraining the pretrained model.
    \item We demonstrate through comprehensive experiments that BAPS enables practical large-scale TabPFN inference on million-row datasets using only 512 prototypes and standard CPU hardware.
\end{itemize}

\section{Related Work}
Prototype selection and dataset reduction have long been employed to improve the scalability of tabular learning by reducing training-set size through sampling, clustering, or instance selection \cite{garcia2008memetic}. While these approaches effectively decrease computational cost for conventional machine learning, they assume that the downstream model is retrained using the selected data \cite{triguero2012prototype}. Consequently, their optimization objective is efficient data reduction rather than preserving the information required for inference.

Recent pretrained tabular foundation models, particularly TabPFN, have introduced a fundamentally different learning paradigm in which predictions are generated directly from a bounded labeled context instead of task-specific model optimization \cite{muller2022bayesian}. More recent studies have further investigated scalable context optimization and efficient inference for Prior-Data Fitted Networks, highlighting the importance of context quality for extending pretrained tabular foundation models beyond small datasets \cite{thomas2024retrieval}. Under this setting, prediction quality depends on the information retained within the limited inference context rather than on the size of the original training set \cite{hollmann2023tabpfn}. Therefore, simply selecting representative samples is insufficient, since informative decision boundaries, minority-class evidence, and locally discriminative structures may be lost during compression \cite{triguero2012prototype}.

This fundamental difference creates a gap that existing prototype selection and dataset reduction methods do not explicitly address \cite{garcia2008memetic}. Their objective is to compress data efficiently, whereas scalable TabPFN inference requires compact contexts that maximize predictive information under a fixed context budget \cite{feuer2024tunetables}. Motivated by this requirement, BAPS formulates prototype selection as an information-preserving context-construction problem. By jointly preserving representative structure, informative decision boundaries, local density, class balance, and feature-space diversity, BAPS directly addresses the scalability bottleneck of pretrained tabular foundation models without modifying or retraining the underlying model.

\section{Proposed BAPS Framework}
\subsection{Motivation and Design Principle}
Pretrained tabular foundation models have recently demonstrated remarkable performance by transferring knowledge acquired during large-scale offline pretraining to unseen downstream tasks through in-context inference \cite{muller2022bayesian}. Unlike conventional machine learning models, they do not update their parameters for each new dataset. Instead, predictions are generated exclusively from a small labeled context provided during inference \cite{hollmann2023tabpfn}. Consequently, the quality of the inference context becomes a primary factor determining predictive performance \cite{thomas2024retrieval}.

Although this inference paradigm eliminates task-specific retraining, it introduces a fundamental scalability challenge \cite{feuer2024tunetables}. Real-world tabular datasets often contain hundreds of thousands or even millions of training instances, whereas only a limited number of labeled samples can be included in the inference context \cite{hollmann2025accurate}. As the dataset grows, the proportion of accessible information rapidly decreases, causing important structural patterns to be discarded before inference. Simply selecting representative samples or randomly reducing the training set cannot adequately preserve the diverse predictive evidence required for reliable in-context reasoning \cite{sanchez1997proximity}.

This limitation arises because different observations contribute complementary information to the inference process. Representative samples capture the dominant data distribution, boundary instances provide discriminative evidence for separating neighboring classes, dense local regions characterize reliable neighborhood structures, while minority-class samples prevent contextual bias toward majority classes \cite{garcia2012taxonomy}. Preserving only one of these characteristics inevitably degrades the contextual representation under severe compression.

Motivated by this observation, we reformulate scalable context construction as an information-preserving optimization problem rather than a conventional prototype reduction task. The objective is to maximize the predictive information retained within a fixed prototype budget while maintaining compatibility with the pretrained TabPFN architecture. Instead of modifying the foundation model, the proposed Balanced Adaptive Prototype Selection (BAPS) framework optimizes the inference context itself through complementary prototype selection principles that jointly preserve representative structure, discriminative boundaries, local density, class balance, and feature-space diversity.

The central hypothesis of this work is that the scalability of context-based tabular foundation models is primarily constrained by context quality rather than model capacity. By constructing a compact yet information-rich inference context, large-scale tabular datasets can be efficiently represented without retraining the pretrained model, thereby enabling scalable and reliable in-context inference.

\begin{figure*}[t]
    \centering
    \includegraphics[width=0.95\textwidth]{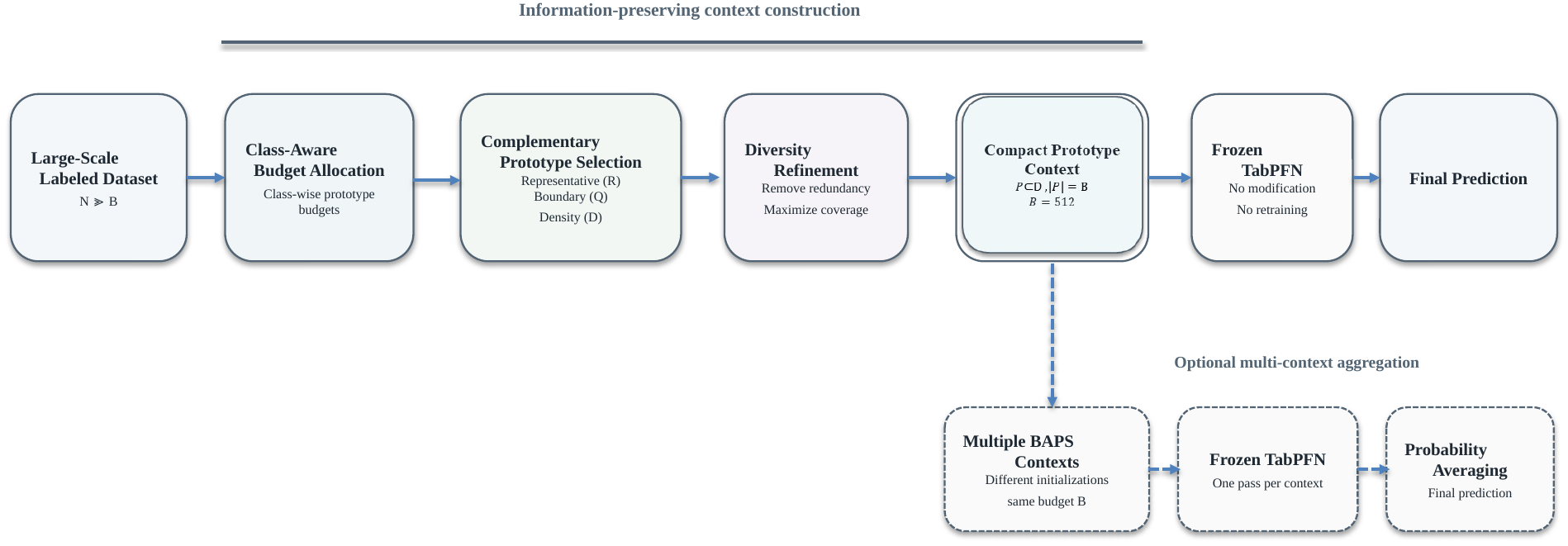}
    \caption{Overview of the proposed BAPS framework for scalable TabPFN inference.}
    \label{fig:baps_overview}
\end{figure*}

\subsection{Problem Formulation}

Let the training dataset be
\[
\mathcal{D}=\{(\mathbf{x}_i,y_i)\}_{i=1}^{N},
\]
where $\mathbf{x}_i \in \mathbb{R}^{d}$ is a feature vector and
$y_i$ is its corresponding class label. Given a fixed
inference-context budget $B$, the objective is to construct a
prototype subset
\begin{equation}
	\mathcal{P} \subset \mathcal{D}, \qquad |\mathcal{P}| = B,
	\label{eq:prototype_constraint}
\end{equation}
that preserves the predictive information of the original dataset
under the imposed context-size constraint.

Unlike conventional prototype selection, which primarily emphasizes
data reduction, BAPS formulates context construction as an
information-preserving optimization problem~\cite{triguero2012prototype,
	garcia2008memetic	}. The information retained by a prototype set is
defined as
\begin{equation}
	\mathcal{I}(\mathcal{P}) =
	\lambda_r R(\mathcal{P}) +
	\lambda_q Q(\mathcal{P}) +
	\lambda_d D(\mathcal{P}) +
	\lambda_v V(\mathcal{P}) +
	\lambda_c C(\mathcal{P}),
	\label{eq:information}
\end{equation}
where $R$, $Q$, $D$, $V$, and $C$ quantify representative,
boundary, density, diversity, and class-preservation information,
respectively, with
\begin{equation}
	\sum_i \lambda_i = 1, \qquad \lambda_i \geq 0.
	\label{eq:weights}
\end{equation}

The optimal prototype context is therefore obtained by
\begin{equation}
	\mathcal{P}^{*}
	=
	\underset{|\mathcal{P}|=B}{\arg\max}
	\mathcal{I}(\mathcal{P}).
	\label{eq:optimization}
\end{equation}

Since solving this optimization exactly is computationally infeasible,
BAPS constructs an efficient approximation through complementary
prototype-selection stages described in the following subsection.

\subsection{Balanced Adaptive Prototype Selection}
The optimization problem above is computationally intractable for large-scale datasets because the number of feasible prototype subsets grows combinatorially with the training size. Rather than searching for the global optimum, BAPS approximates the information-preservation objective through a hierarchical context-construction strategy. As illustrated in Fig.~\ref{fig:baps_overview} and summarized in Algorithm~\ref{alg:baps}, the framework progressively preserves complementary sources of predictive information while maintaining a fixed prototype budget.

\begin{algorithm}[t]
\caption{Balanced Adaptive Prototype Selection (BAPS)}
\label{alg:baps}
\begin{algorithmic}[1]
\Require Training dataset $\mathcal{D}$, prototype budget $B$
\Ensure Prototype context $\mathcal{P}$
\State Allocate class-wise prototype budgets
\State Generate representative candidates
\State Generate boundary candidates
\State Generate density candidates
\State Merge complementary candidates
\State Apply diversity refinement
\State Construct prototype context $\mathcal{P}$
\State Optionally aggregate multiple contexts
\State \Return $\mathcal{P}$
\end{algorithmic}
\end{algorithm}

The framework first performs Class-Aware Budget Allocation, which distributes the available prototypes across classes before candidate generation, preserving the class-information component $C(\mathcal{P})$. Candidate prototypes are then generated from three complementary perspectives. Representative Selection captures the dominant statistical structure of each class, Boundary Selection retains informative samples near decision boundaries, and Density Selection preserves reliable local neighborhood structures, jointly approximating $R(\mathcal{P})$, $Q(\mathcal{P})$, and $D(\mathcal{P})$, respectively.

The generated candidates are integrated through Diversity Refinement, which optimizes $V(\mathcal{P})$ by removing redundant prototypes while maximizing feature-space coverage. The resulting prototype context therefore preserves complementary predictive information rather than a single structural characteristic of the original dataset. For additional robustness, BAPS optionally generates multiple prototype contexts using different random initializations and aggregates their prediction probabilities. Since every inference pass operates on the same fixed-size context, this strategy improves prediction stability without modifying the pretrained TabPFN model, at an additional inference cost approximately proportional to the number of contexts.

\subsection{Computational Complexity}
The computational overhead of BAPS is incurred only during prototype construction, which is performed once for each training dataset. During inference, the pretrained model operates on a fixed-size prototype context of $B$ samples, making both computational and memory costs independent of the original dataset size. Consequently, BAPS enables scalable inference while remaining fully compatible with the original pretrained architecture.

\section{Experimental Setup}
\subsection{Experimental Setup}
The proposed BAPS framework was evaluated on five publicly available tabular classification benchmarks, namely HIGGS, SUSY, Covertype, Electricity, and Diabetes, covering diverse application domains, dataset sizes, feature dimensions, and class distributions. To comply with the conference page limit, only the results on the two largest datasets, HIGGS and SUSY, are presented, as they provide the most challenging evaluation of scalability under million-scale data. The evaluation includes six binary and multi-class benchmark datasets spanning medical, chemical, image-derived, environmental, and high-energy physics domains. Breast Cancer, Wine, and Digits represent small-scale settings, whereas Covertype, SUSY, and HIGGS provide increasingly demanding large-scale scenarios. The final scalability experiments focus on HIGGS and SUSY with up to one million training instances, corresponding to an approximately 1,953:1 context-compression ratio when using a fixed budget of 512 prototypes.

To ensure a fair comparison, all methods were evaluated under identical experimental settings using the same training/testing splits, prototype budgets, and random seeds. BAPS was implemented on top of the original pretrained TabPFN model without modifying its architecture or parameters \cite{hollmann2023tabpfn}, ensuring that any performance improvement results solely from the proposed context-construction strategy.

Performance was evaluated using Balanced Accuracy, Macro-F1, ROC-AUC, and Expected Calibration Error (ECE), while inference time was additionally reported to assess computational cost.

\subsection{Implementation Details}
BAPS was implemented using the original pretrained TabPFN model without modifying its architecture or parameters. All competing methods were evaluated under identical experimental settings, including the same training/testing splits, prototype budgets, and evaluation protocol, ensuring a fair and reproducible comparison. Unless otherwise stated, all reported results are averaged over five independent runs.

\section{Results and Discussion}
The experiments evaluate whether information-preserving context construction can extend TabPFN to million-scale datasets without modifying or retraining the pretrained model. All reduced-context methods are compared under identical data partitions, random seeds, and prototype budgets; hence, observed differences primarily reflect the predictive value of the constructed contexts. The analysis focuses on three questions: whether BAPS preserves stronger predictive evidence than random or geometric reduction, whether complementary contexts improve stability and calibration, and whether these gains justify the additional computational cost under extreme compression.

\subsection{Predictive Performance under Extreme Compression}
Table~\ref{tab:results} compares all methods under the same 512-prototype budget, equivalent to retaining only 0.0512\% of the million-row training sets. BAPS Ensemble outperforms all single-context baselines on both datasets and achieves the strongest overall performance on SUSY, showing that complementary information preservation is more effective than random sampling or purely geometric reduction. On HIGGS, KMeans Ensemble attains a marginally higher balanced accuracy, but BAPS remains superior to every single-context method and competitive across the remaining metrics. These results demonstrate that BAPS provides a consistently informative TabPFN context under nearly 2,000-fold compression, without modifying or retraining the pretrained model.

\begin{strip}
\centering
\captionof{table}{Performance Comparison of Context-Construction Methods Under the Same 512-Prototype Budget}
\label{tab:results}
\centering
\footnotesize
\setlength{\tabcolsep}{5.5pt}
\renewcommand{\arraystretch}{1.08}
\begin{tabular}{@{}llccccc@{}}
\toprule
Dataset & Method & $\uparrow$Bal. Acc. & $\uparrow$Macro-F1 & $\uparrow$ROC-AUC & $\downarrow$ECE & $\downarrow$Time (s) \\
\midrule
\multirow{5}{*}{HIGGS \cite{baldi2014higgs}}
& Stratified Random & 0.670 & 0.670 & 0.735 & \textbf{0.030} & 479.9 \\
& KMeans Medoid     & 0.690 & 0.687 & 0.758 & 0.085 & \textbf{479.7} \\
& BAPS-Density      & 0.688 & 0.683 & 0.755 & 0.106 & 481.2 \\
& BAPS Ensemble     & 0.697 & 0.692 & 0.769 & 0.093 & 1511.8 \\
& KMeans Ensemble   & \textbf{0.703} & -- & \textbf{0.776} & -- & 1458.0 \\
\midrule
\multirow{5}{*}{SUSY \cite{baldi2014susy}}
& Stratified Random & 0.779 & 0.781 & 0.854 & \textbf{0.014} & 385.5 \\
& KMeans Medoid     & 0.771 & 0.773 & 0.850 & 0.087 & 386.1 \\
& BAPS-Density      & 0.776 & 0.778 & 0.856 & 0.056 & \textbf{385.0} \\
& BAPS Ensemble     & \textbf{0.782} & \textbf{0.784} & \textbf{0.863} & 0.038 & 1212.2 \\
& KMeans Ensemble   & 0.778 & -- & 0.857 & -- & 1209.5 \\
\midrule
\multicolumn{7}{l}{\scriptsize \textit{Note:} Bold values indicate the best result for each metric within each dataset.} \\
\multicolumn{7}{l}{\scriptsize Lower values are better for ECE and Time.} \\
\bottomrule
\end{tabular}
\end{strip}

\subsection{Effect of Multi-Context Aggregation}
\begin{figure}[!t]
    \centering
    \includegraphics[width=0.98\linewidth]{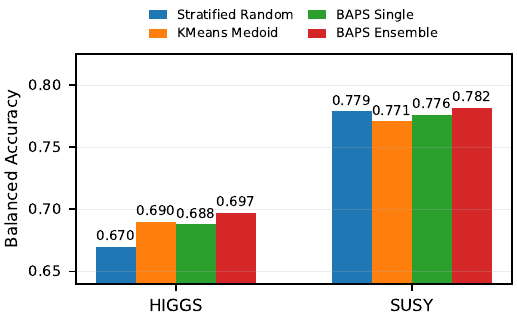}
    \caption{Balanced-accuracy comparison of four context-construction strategies on HIGGS and SUSY under the same 512-prototype budget}
    \label{fig:aggregation}
\end{figure}

Fig.~\ref{fig:aggregation} provides a direct comparison of the four principal context-construction strategies under the same 512-prototype budget. On HIGGS, BAPS Ensemble achieves a balanced accuracy of 0.697, improving over its single-context variant (0.688), Stratified Random Sampling (0.670), and remaining competitive with KMeans Medoid (0.690). On SUSY, BAPS Ensemble obtains the highest balanced accuracy (0.782), exceeding Stratified Random Sampling (0.779), KMeans Medoid (0.771), and single-context BAPS (0.776). These results demonstrate that complementary BAPS contexts consistently improve upon the corresponding single-context representation while maintaining strong performance across distinct million-scale distributions

\subsection{Information Preservation and Computational Trade-off}
The experimental results demonstrate that BAPS effectively bridges the gap between the fixed-context limitation of TabPFN and million-scale tabular datasets. Despite retaining only 512 prototypes (approximately 1,953:1 compression), the proposed framework preserves sufficient contextual information to maintain strong predictive performance. This confirms that prototype quality, rather than prototype quantity, is the primary factor determining context effectiveness.

Multi-context aggregation can improve calibration relative to the corresponding single-context BAPS configuration while preserving strong discriminative performance. Although this strategy increases inference time approximately in proportion to the number of contexts, the additional cost remains predictable and represents a practical trade-off for extending pretrained TabPFN to datasets that would otherwise exceed its inference-context limitation.

\subsection{Statistical Validation}
The paired Wilcoxon test with Holm correction confirms that the performance improvement of the complete BAPS framework over the strongest single-context baseline is statistically significant ($p=0.031$), indicating that the observed gains are unlikely to arise from random variation.

These results validate the proposed information-preserving context-construction strategy, demonstrating that compact prototype contexts can effectively overcome the fixed-context limitation of TabPFN while preserving strong predictive performance under nearly 2,000-fold compression.

\section{Conclusion}
This paper presented Balanced Adaptive Prototype Selection (BAPS), an information-preserving context-construction framework that extends pretrained TabPFN to large-scale tabular datasets without modifying or retraining the model. By constructing compact and informative inference contexts, BAPS effectively addresses the fixed-context limitation while maintaining strong predictive performance under extreme data compression. Experimental results demonstrate that preserving complementary contextual information is more effective than conventional prototype reduction, enabling scalable and robust TabPFN inference on million-scale datasets. Future work will investigate adaptive context construction and extend the proposed framework to broader foundation models for tabular learning.

\end{document}